# Fast Dust Sand Image Enhancement Based on Color Correction and New Membership Function

**Ali Hakem Alsaeedi[1,2], Suha Mohammed Hadi[1], Yarub Alazzawi[3]**
[1]Informatics Institute for Postgraduate Studies, Iraqi Commission for Computer and Informatics, Bagdad, Iraq
[2]College of Computer Science and Information Technology, University of Al-Qadisiyah, Al Diwaniyah, Iraq
[3]Al-Khwarizmi College of Engineering, University of Baghdad, Baghdad, Iraq
corresponding author's e-mail: phd202130687@iips.icci.edu.iq

**Abstract**. Images captured in dusty environments suffering from poor visibility and quality. Enhancement of these images such as sand dust images plays a critical role in various atmospheric optics applications. In this work, proposed a new model based on Color Correction and new membership function to enhance san dust images. The proposed model consists of three phases: correction of color shift, removal of haze, and enhancement of contrast and brightness. The color shift is corrected using a new membership function to adjust the values of U and V in the YUV color space. The Adaptive Dark Channel Prior (A-DCP) is used for haze removal. The stretching contrast and improving image brightness are based on Contrast Limited Adaptive Histogram Equalization (CLAHE). The proposed model tests and evaluates through many real sand dust images. The experimental results show that the proposed solution is outperformed the current studies in terms of effectively removing the red and yellow cast and provides high quality and quantity dust images.

## 1. Introduction

Sandstorms have hit several countries, and their extent and intensity are growing [1]. When taking images or videos in dusty sand weather, there are usually various impairments in the quality of vision and detection of objects[2]. Generally, the image appears dimmed in dusty sand and has low color contrast, poor visibility, and new high color tones of yellowish or reddish[3]–[8]. Technically, the low image quality in terrible atmospheric sandstorms is because the sand dust particles scatter and absorb a specific light spectrum. Therefore, the corruption in sand dust images will significantly impact various atmospheric optics applications, typically working outdoors during manifestation weather conditions. Such as video surveillance for public safety services[9], intelligent transportation systems(ITS) [10], video tracking and monitoring systems for vehicles[11], Remount sensing[12], and so on. For computer vision applications, developing a sand-dust image restoration approach is urgent to enhance the quality and vision of the image[13].

Uneven absorption of color spectrums of the light reflected by the object leads to distortion of the image that reaches the camera[5]. The red, green, and blue wavelengths (λ) are 630 µm, 532 µm, and 465 µm, respectively. According to [14], there is an inverse relationship between atmospheric particle size and wavelength. In underwater or haze, the absorption of red is more than green and blue channels; therefore, the image, in this case, seems greenish or bluish. The radius of the water drop is about 14 µm, the haze is 10-2-1 μm, and the sand particle size is about 100 μm[15]: the image degradation, veil and color histogram between sand-dust image and mist, fog, haze or underwater image. Therefore, Dehazing and underwater image techniques will fail to improve sand-dust images. Figure 1 illustrates example of images impact by fog (a), two underwater images(b, and c), and two sand-dust images (d, and e) with shown the atmospheric veils and histograms of red,green, and blue channel of each iamge.

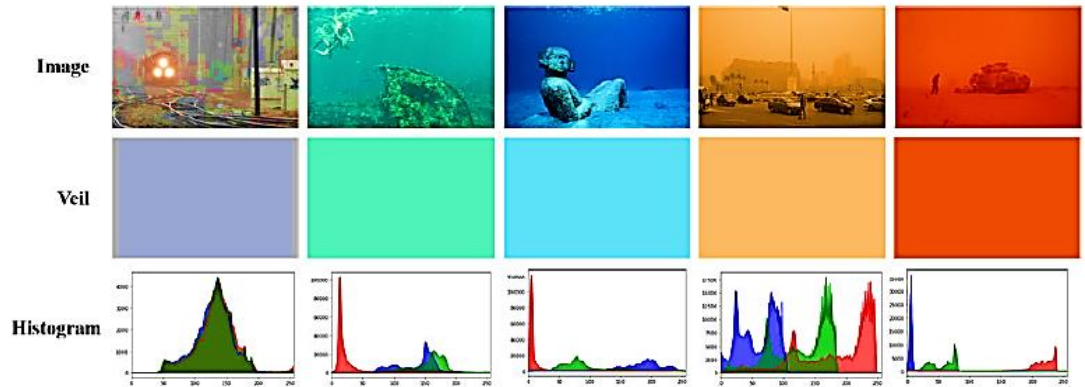

Figure 1: Veil and histogram of degraded image underwater and outdoor [16]

Figure 1 shows that the image's veil relies on the color's intensity. Depending on the sand's strength, images in sand dust are yellowish or reddish. Therefore, color balancing and correction are necessary for enhancing the sand-dusty image[5], [17], [18][19].
Several studies have proposed various image processing techniques to develop the approach of sand dust imaging.
Shi et al.[20] proposed a hybrid model for enhancing the sand-dust image based on two techniques DCP and Lab color space. The proposed model traded the halo in the sand-dust image by DCP, while the correct shifting in the color of the final appearance was based on an analysis of the LAB color space. The lab color space model is stretched to improve image brightness and contrast. This method reduced the effect of the yellow tone in sand dust images. The limitation of the proposed model the recovered image was too dim.
In [21], Eigenvalue is used for each color (R, G, and B) to normalise (balance) the color contrast. Finally, the adaptive dark channel prior (ADCP) is applied for color correction. This method allows images with a lower degree of sand to be quickly enhanced; the visual effect is limited with severe degradation.
Gao et al. [22] proposed a new multiscale of the retinex to process the Y component. It reduced the influence of stray light in the air and improved image clarity. The YUV color space of an image processing implementation can provide better subjective image quality than the RGB color space; therefore, using the YUV color space causes the proposed model to improve contrast and show more detail. The system is fast and potentially suitable for real-time image processing applications. However, this approach could not recover degraded images with significant sand dust.
Al- Ameen [23] proposed a fuzzy model for improving sand-dust images based on developed fuzzy intensification to be suitable for enhancing sand dust images. This method corrects the color based on the three thresholds of the color channel. The proposed model is efficient in restoring the color channel. However, the tuning method uses a constant value that causes a halo effect similar to color distortion; therefore, this method is unsuitable for various images.

## 2. Proposed model

The proposed model consists of three steps: Color correction, haze removal, and contrast enhancement
- Step 1: Shift color correction based on membership function.
- Step 2: Remove haze.
- Step 3: contrast and enhance detail.

*2.1. Correcting of Shift color:*

Most of the image information in the YUV color space is stored in the Y component. In the proposed model, we only adjust the two components separately, U and V, to not destroy image visibility. In the YUV color space, the UV color component is used for the first time to remove the color cast[22]. Equations 1, 2, and 3 are used to convert the color space of the sand dust image from RGB to YUV.

$$Y = 0.299\,R + 0.587\,G + 0.114B \quad (1)$$
$$U = -0.168736\,R - 0.331264\,G + 0.5\,B \quad (2)$$
$$V = 0.5\,R - 0.418688\,G - 0.081312\,B \quad (3)$$

R (red value), G (green value), and B (blue value) are components of the RGB. After extracting Y, U and V, both U and V values are adjusted in the proposed model according to the proposed new membership function shown in equations 4 and 5.

$$U'_i = U_i - \frac{1}{m \times n} \sum_{j=0}^{n} \sum_{k=0}^{n} U_{i,k} \quad (4)$$
$$V'_i = V_i - \frac{1}{m \times n} \sum_{j=0}^{n} \sum_{k=0}^{n} V_{i,k} \quad (5)$$

Where: $U_i$, $V_i$ are the value of the U and V components of YUV color space of a pixel in position j,k. The m, n is the dimension of the image. After adjusting the U and V, the YUV is retransformation into RGB color space based on equations 6,7, and 8[22]

$$R = Y + 1.402\,V \quad (6)$$
$$G = Y - 0.3456U - 0.7145 * V \quad (7)$$
$$B = Y + 1.7710 * U \quad (8)$$

Figure 3 shows the application of shift color correction on image a to obtain image b

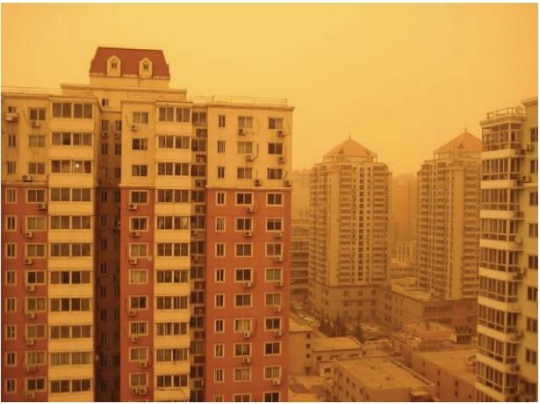 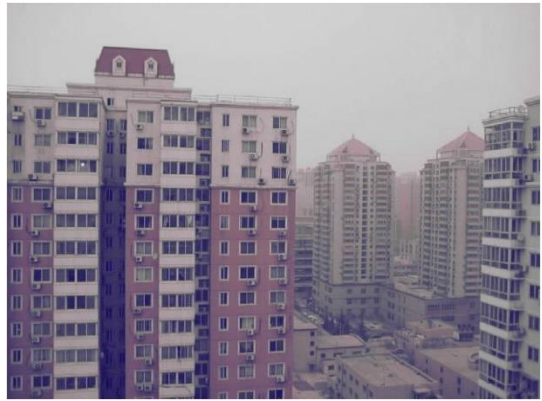

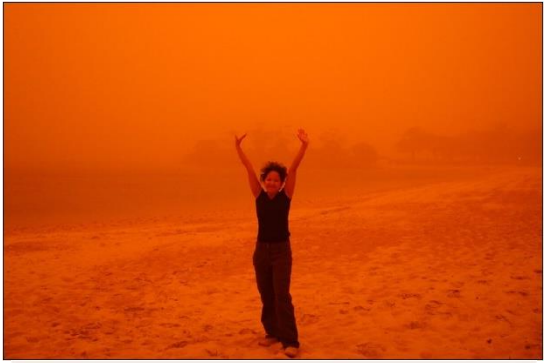 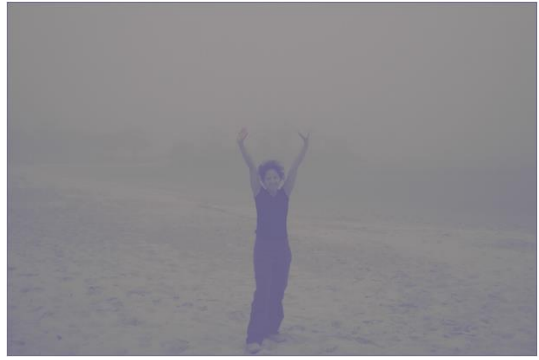

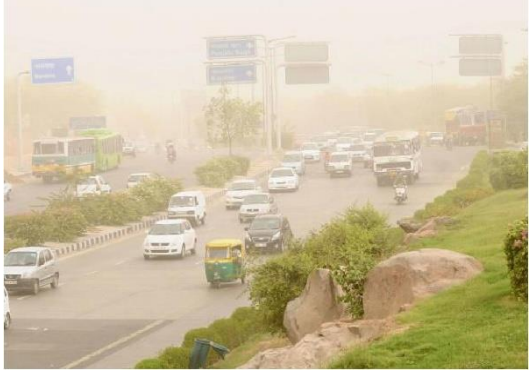 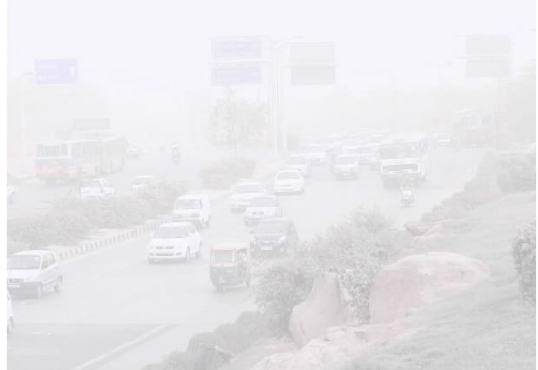

          a: original image                          b: image after correct color shift

Figure 3: Example of application of step1 (shift color correction)

## 2.2. Remove haze

The Adaptive A-DCP eliminates light influence to improve image clarity. The output image of shifting color correction is used as input to the A-DCP. The image restoration based on a physical model enhances the image according to the analysis of the physical characteristics of pixels. Image restoration operations help to remove noise and blur from the image, either linear or nonlinear. Figure 4 the application of removing haze by A-DCP on image a to obtain image c.

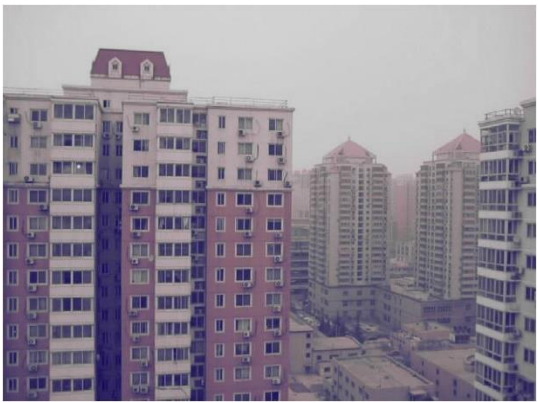 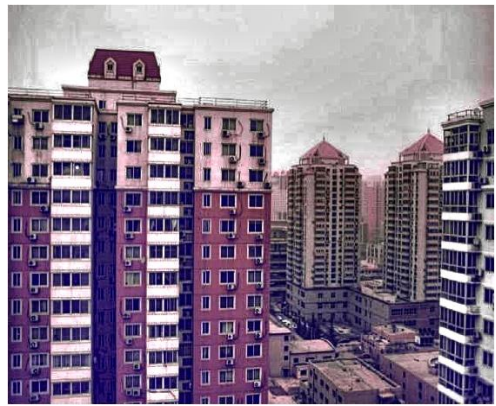

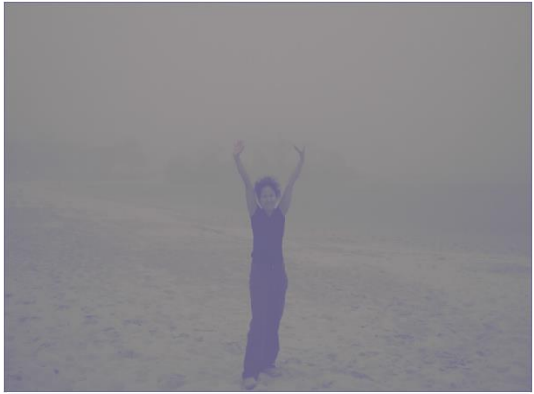
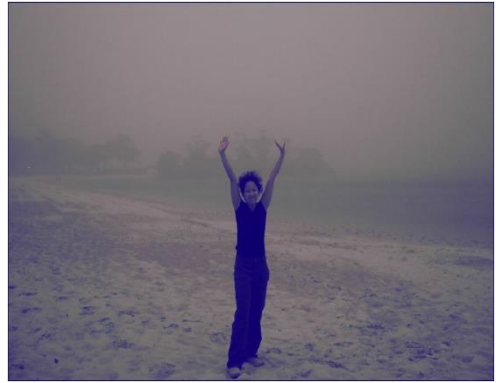
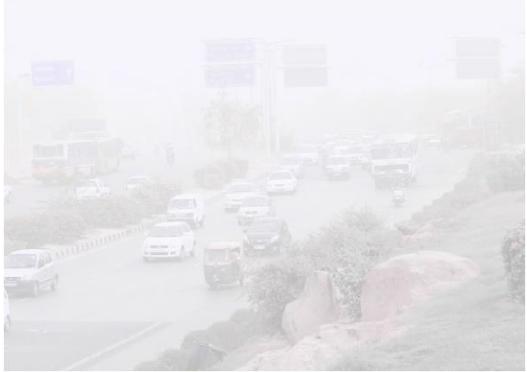
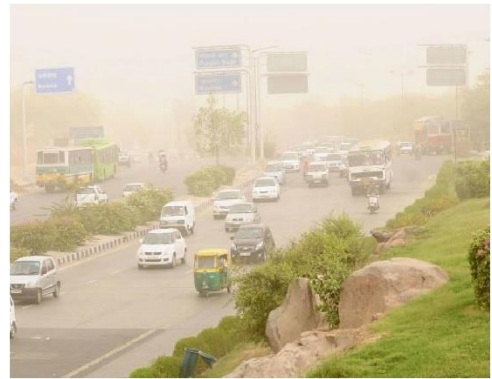

      b: image after correct color shift        c: image after removing haze

Figure 4: Example of application of step2(Remove haze)

### 2.3. Contrast and enhance detail

The last step in the proposed system includes enhancing the vision and improving the details of the image. According to [22], Contrast Limited Adaptive Histogram Equalization (CLAHE) improves contrast and enhances detail. It may boost contrast and brightness, enhance local information, and be suitable for real-time demands. Figure5 the application of contrast and enhanced detail by CLAHE on image b to obtain image d.

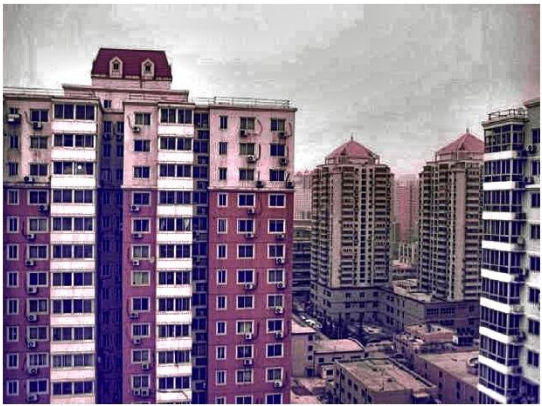
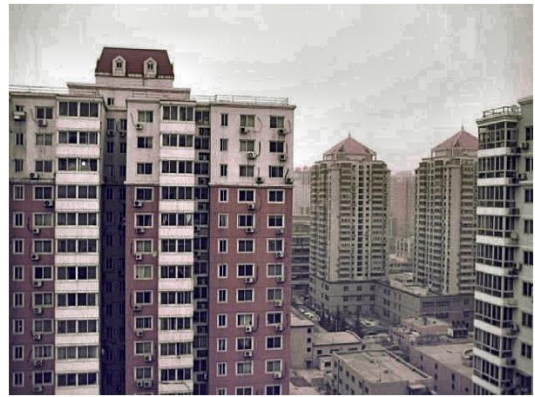

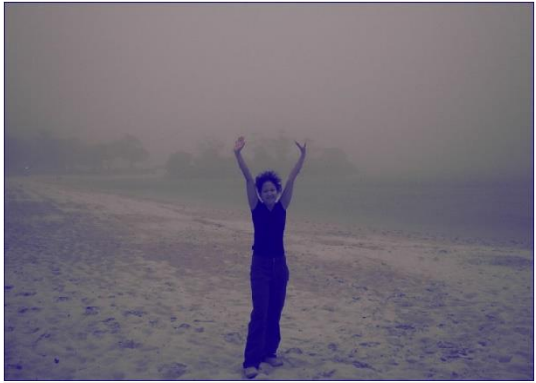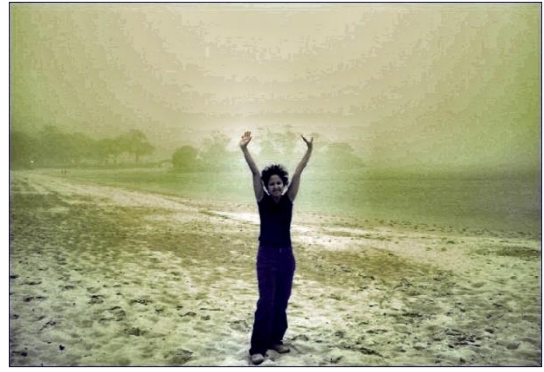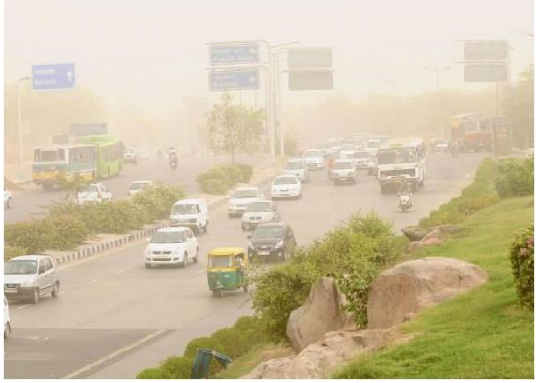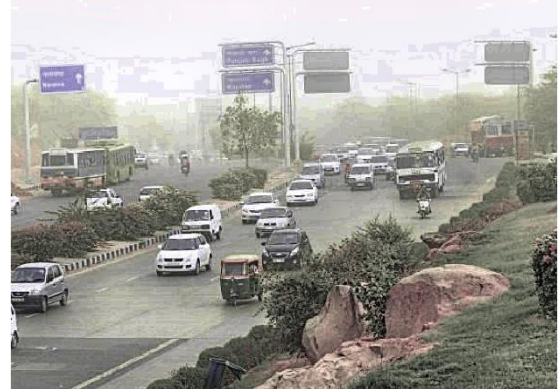

    c: image after removing haze      d: image after removing contrast and enhanced detail

. Figure 4: Example of application of step3(Remove contrast and enhanced detail)

### 3. Experimental Results and Discussion

The distorted sand dust image is balanced by the proposed method. Since the balanced image has similar characteristics to an out-of-focus image, the decolorization algorithm is applied using the proposed ADCP method. In this section, the suitability of the proposed method for enhancing sand dust images is demonstrated subjectively and objectively. The proposed method and state-of-the-art methods are compared subjectively and objectively. In addition, the adverse weather detection dataset [16] is used to compare their application in different circumstances. The dataset in [16] consists of 245 sand dust images. The performance of our approach is compared with the performance of two state-of-the-art methods ([23] *and* [24]).

*1.1. Objective Comparison*

The term No-reference Image Quality Assessment (NR-IQA) in image evaluation is measuring the quality image without depending on the reference of the image[25]. Digital images degrade during storage, compression, transmission, or bad weather [26],[27]. The most reliable method of evaluating image quality is to ask people to give their opinions on a series of test images. However, this is an expensive and time-consuming procedure that cannot be used in real-time systems. In the case of non-reference images, it is not easy to obtain a reference image to evaluate the recovered image[28]. In our case, we computed three coefficients, factors e and σ, and $\bar{r}$ [29]. The values of the coefficients depend on the number of visible edges, and the coefficient depends on the contrast value of visible edges. Equations 9,10, and 11 compute the coefficients e and σ, respectively.

$$e = \frac{n_r - n_o}{n_o} \tag{9}$$

Where: $n_0$, $n_r$ the number of visible edges in the original and restored image, respectively.

$$\bar{r} = \exp\left[\frac{1}{n_r}\sum \log r_i\right] \tag{10}$$

Where $r_i$ the contrast value of edge

$$\sigma = \frac{n_s}{\dim_x \times \dim_y} \tag{11}$$

Number $n_s$ of pixels that are saturated (black or white).
When comparing two images, the image that achieves the highest value of is better.

Table 1 compares the proposed model and two state-of-the-art models, [23] and [24], according to e, σ, and r¯ calculated for sand dust images restored. The result in table 1 approved that the proposed model achieved performance better than [23] and [24]. The time performed by Al-Ameen [23] is the lowest among the systems, but with somewhat not good results. In[23], the author relied on training the system on fixed cases, the limit of problems taken as a limitation for his work.

**Table 1.** Comparison of the e and σ, and $\bar{r}$ score of the images in [17] (If the score is high, the image is enhanced well).

| Model | E | $\bar{r}$ |  | Time(sec) |
|---|---|---|---|---|
| **Al-Ameen** [23] | 1.6 | 1.76 | 0.25 | 1.43 |
| *Shi et al.* [24] | 3.85 | 3.13 | 0.29 | 2.76 |
| **Proposed model** | 5.43 | 3.77 | 0.33 | 1.58 |

*1.2. Qualitative Comparison*

Figure 5 shows the image recovery by the proposed method and Al-Ameen [23], Shi et al. [24] on different sandstorm cases. The image was restored by [23], and the model could remove the unwanted color casts, but the restored image is dark, and the details are lost. The authors in [24] improved the image contrast, but the contrast is increased too much, and the restoration results are highly distorted. Our model restored results are more natural in color, clearer in detail and approximately similar to real images.

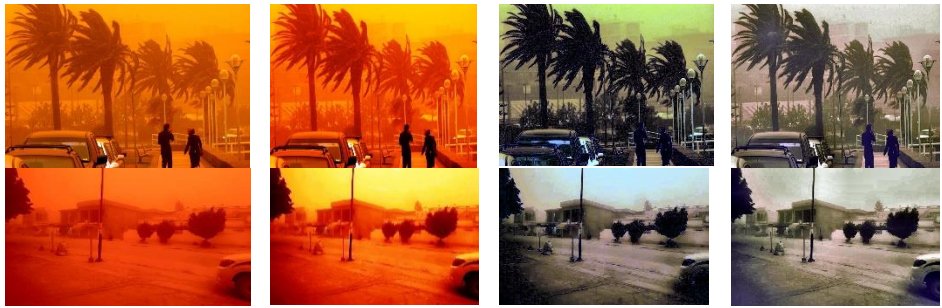

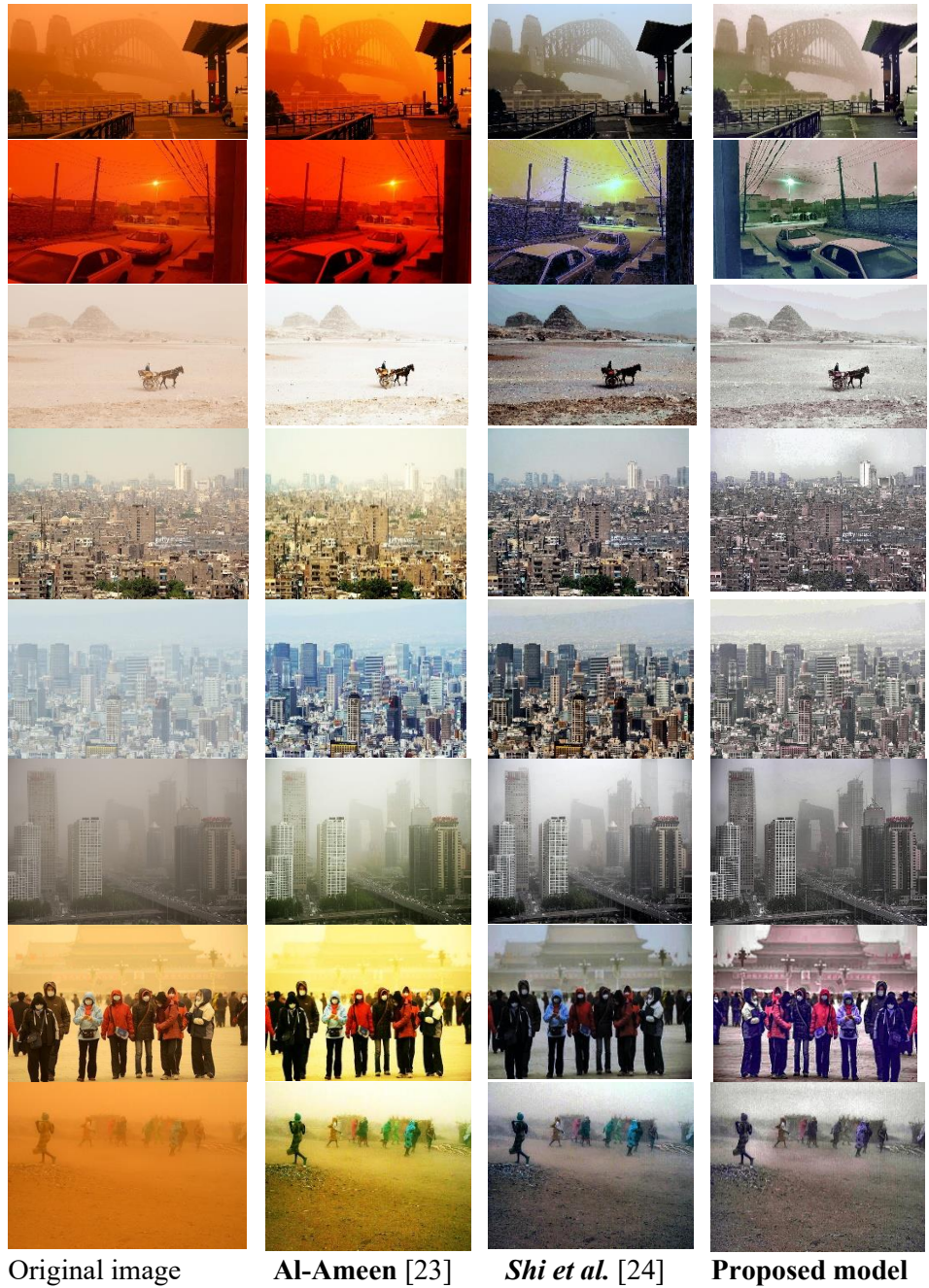

Figure 5: Comparing sandstorm images based on qualitative: Al-Ameen [23] Shi et al. [24], and Proposed model

## 4. Conclusion

In this paper presents a method based on Shift Colour Correctio, Dark Channel Prior (DCP) and Contrast Limited Adaptive Histogram Equalisation (CLAHE) to improve the sand dust image. In YUV space, the UV color component removes the color cast for the first time. Most of the image information in the YUV color space is stored in the y component. Therefore, the separate adjustment of the two components, U and V, reduces the image's distortion. Dust removal with an adaptive DCP equalisation method. Contrast stretching and image brightness enhancement based on CLAHE. Images taken in a dusty environment suffer significantly from poor visibility and quality. The enhancement of these images plays a vital role in various atmospheric optics applications. Restoring sand dust images is as important as removing haze and enhancing underwater images. The proposed method can effectively remove the yellow cast and dust haze effect and provides typical visual color and a clear image. Experiments with many real sand dust images show that the proposed method can achieve good color fidelity and reasonable brightness. The limitation of the proposed model is that the images recovered by the system are somewhat bluish because the color shift correction is applied only to U and V. The color shift correction is applied only to the U and V components. The Y component of YUV remains unadjusted, resulting in a deterioration of the bluish cast of some images. For future work, we suggest shifting color correction based on the three components of the YUV color space of the sand dust image.